\documentclass[runningheads]{llncs}

\usepackage{eccv}

\usepackage{eccvabbrv}

\usepackage{graphicx}
\usepackage{booktabs}

\usepackage[accsupp]{axessibility}  

\usepackage[pagebackref,breaklinks,colorlinks,citecolor=eccvblue]{hyperref}

\usepackage{orcidlink}
\begin{document}

\title{FSViewFusion: Few-Shots View 
Generation of Novel Objects} 

\titlerunning{FSViewFusion}

\author{Rukhshanda Hussain\inst{1}
Hui Xian Grace Lim\inst{1} Borchun Chen\inst{2} Mubarak Shah \inst{1} 
Ser Nam Lim\inst{1}}

\authorrunning{Hussain et al.}

\institute{$^1$University of Central Florida 
$^2$University of Maryland}
\maketitle

\begin{abstract}
  Novel view synthesis has observed tremendous developments since the arrival of NeRFs. However, Nerf models overfit on a single scene, lacking generalization to out of distribution objects. Recently, diffusion models have exhibited remarkable performance on introducing generalization in view synthesis. Inspired by these advancements, we explore the capabilities of a pretrained stable diffusion model for view synthesis without explicit 3D priors. Specifically, we base our method on a personalized text to image model, Dreambooth~\cite{ruiz2023dreambooth}, given its strong ability to adapt to specific novel objects with a few shots. Our research reveals two interesting findings. First, we observe that Dreambooth can learn the high level concept of a view, compared to arguably more complex strategies which involve finetuning diffusions on large amounts of multi-view data. Second, we establish that the concept of a view can be disentangled and transferred to a novel object irrespective of the original object's identify from which the views are learnt. Motivated by this, we introduce a learning strategy, FSViewFusion, which inherits a specific view through only one image sample of a single scene, and transfers the knowledge to a novel object, learnt from few shots, using low rank adapters. Through extensive experiments we demonstrate that our method, albeit simple, is efficient in generating reliable view samples for in the wild images. Code and models will be released.

\end{abstract}

\section{Introduction}\label{sec:intro}
Novel view synthesis has always been a fundamental problem in 3D vision, with its applications extending to several engineering problems like video enhancement, virtual reality, 3D content creation, etc. Implicit neural representations as in NeRF \cite{mildenhall2021nerf} and approaches based on it like \cite{barron2021mip}, \cite{yu2021pixelnerf}, \cite{deng2022depth}, \cite{niemeyer2022regnerf} have demonstrated great success in reconstructing 2D views of a scene given the camera positions for a view. However, these methods even while addressing sparse view reconstructions are often restricted by the need to procure multi-view data for the respective scene to be reconstructed. Recently, denoising based diffusion models \cite{ho2020denoising}, \cite{song2020denoising}, \cite{rombach2022high} have been extensively used in the task of novel view synthesis because of its generalizability. However, these methods \cite{gu2023nerfdiff}, \cite{tseng2023consistent}, \cite{ye2023consistent}, often rely on three dimensional priors to condition their generative models, which often require extrinsic and intrinsic camera poses for reliable view reconstructions.

\begin{figure*}[!h]
    \centering
    \includegraphics[width=\linewidth]{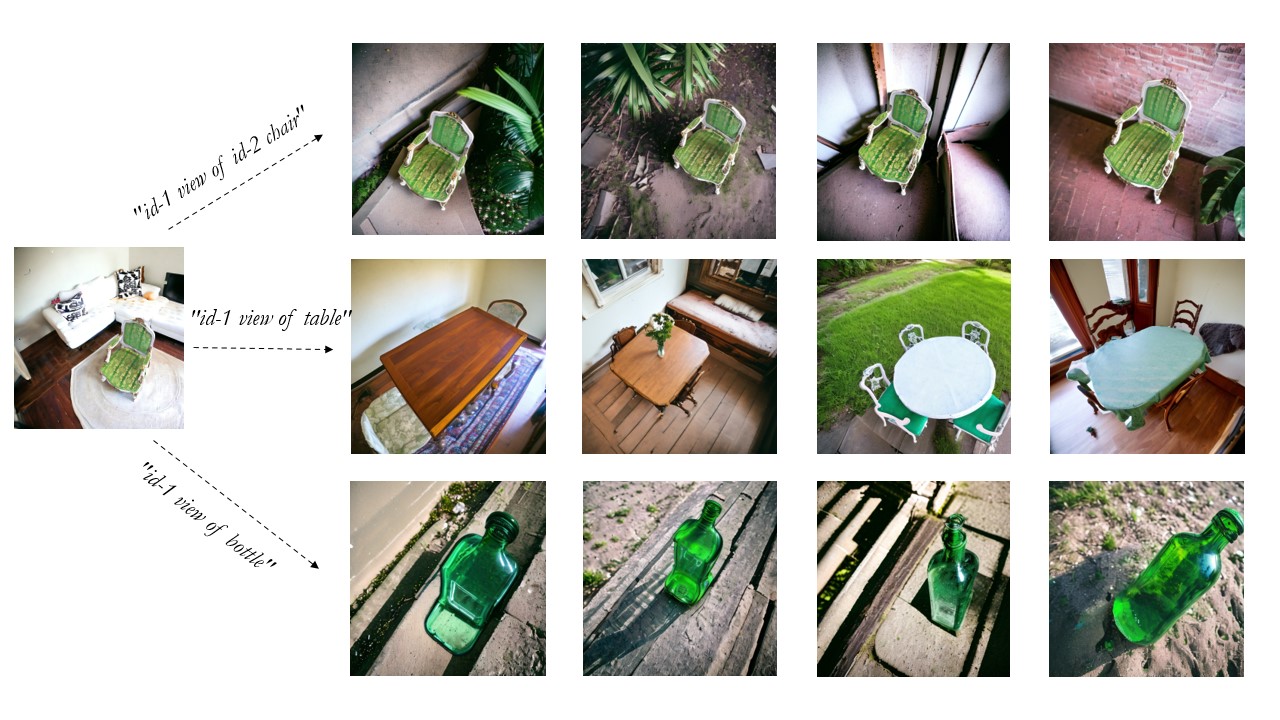}
    \caption{\textbf{Impact of transferring a view concept to different object distributions in diffusion's pretrained space.} Row 1 shows that a particular view of a chair can be retrieved with an unique id assigned for the concept. Row 2 and 3 show that the abstract concept of this view is learnt in disentangled manner which allows reliable reconstruction of other object in the same view point.}
    \label{motivation}
\end{figure*}

Interestingly, human brain does not require any camera poses to perceive a viewpoint of an object in a real world setting. It can be analogous to a model which has been trained on a vast different instances of data which allows it to provide an estimate of the view. Considering this fact, we question whether pretrained diffusion models trained on a virtually exhaustive amount of data is already capable of understanding the viewpoint of any object through its spatial relations to other objects in background by visual cue alone? In order to answer this, we conduct a simple experiment with text to image personalised models (in our case DreamBooth \cite{ruiz2023dreambooth}).
Utilizing Dreambooth's setup is appealing for two reasons. First, DreamBooth produces high fidelity to the subject context given with as few as 3-4 samples of the context. Second, we can still leverage the massive pretraining of the underlying diffusion model to generate new concepts. We generate a few samples of several views of chair with different backgrounds synthetically through in-painting \cite{yu2023inpaint} and assign a unique id for each view. We observe that not only does the unique id binds faithfully to the context of view but also it can reliably disentangle the context to a random object distribution in diffusion model's knowledge space as shown in \autoref{motivation}. 

Motivated by these findings, we hypothesize that diffusion model are capable of learning high level concepts such as view analogous to the concept of style \cite{wang2023styleadapter} \cite{sohn2024styledrop} \cite{shah2023ziplora}, gender \cite{wang2024concept} etc in a disentangled manner from the object identity which shares the concept. Therefore, we rethink the problem of novel view synthesis as this: 
given the knowledge of a few views in the form of 2D images for a single object, a pre-trained diffusion model learns the views as a high level concept, and can transfer the knowledge of the views in generating novel views of a new object.
Further, \autoref{motivation} reveals that object pose and camera view are not the same thing that has been ambiguous in the current literature. The camera view of two objects can be the same but their object poses might change altogether. 

Given that we are not provided with any 3D knowledge of the camera/object pose under this setting, we consider for exposition the camera view as a visual concept where we imagine that the camera is placed at some location in a 3D scene focused on the object. Now, we imagine removing this object in the 3D scene, replacing it with a novel object. The camera position remains unchanged while the degree of freedom is that the pose of the novel object can be different. Indeed, leveraging the findings from Dreambooth, it is conceivable to learn the aforementioned view/scene concept as well as the novel object concept separately using a minimal amount of 2D data with no associated pose meta data, which is easy to procure. By combining the two resulting models, we show that the view and scene can now be transferred to the novel object. 

Our contributions can be summarized as follows: 
\begin{enumerate}
    \item We provide empirical evidence that a view, which describes the spatial relationship of an object with its surrounding in a 3D space, can be treated as a concept to train contextually personalised diffusion models.
    \item We design a learning strategy which first learns the concept of view, second learns the user specified object and finally merges the two concepts to generate novel views of out-of-distribution images. To avoid catastrophic forgetting, the view, object and the final merged concepts are all learned with LoRA \cite{hu2021lora}.
    \item Tapping into the few-shots nature of Dreambooth, our method operates under a few-shots constraint, using as few as a single sample of an object to learn a view LoRA and 3-4 samples to learn a novel object LoRA, thereby avoiding extensive pretraining on multi-view data. We call our method FSViewFusion (FS = Few-Shots).
    \item We provide extensive ablations of FSViewFusion for several uses cases using in the wild images and benchmark our method on widely used DTU dataset for the task of novel view synthesis. 
\end{enumerate}

\section{Related Works}
\subsection{3D Generative Models}
Text to image models like DALLE\cite{ramesh2022hierarchical}, Latent Diffusion \cite{rombach2022high}, Imagen \cite{saharia2022photorealistic} have demonstrated incredible performance in generating high resolution, realistic images in a zero shot manner. Dreamfusion \cite{poole2022dreamfusion} and its follow up works like Prolific Dreamer \cite{wang2023prolificdreamer} are extension of text to image models to 3D by introducing some 3D aware component to the methods. Unlike text to 3D models, view synthesis from a few images have the constraint of conserving the visual features of the input image. Methods like DiffRF \cite{muller2023diffrf} directly noise and de-noise a radiance field followed by volume rendering to obtain realistic object centric views. But these methods require ground truth radiance fields in order to train. Furthermore, since processing radiance fields is computationally expensive, it cannot generate high resolution views. Camera pose conditioned diffusion models like 3DIM \cite{watson2022novel} train a diffusion network from scratch to generate novel views. However, training a network from scratch fails to utilise the information locked in Stable diffusion models which are already trained on millions of images.

\subsection{Novel View Synthesis using Implicit Neural Fields}
The domain of novel view synthesis, in recent times, have largely grown around the implicit neural radiance fields (NeRFs) \cite{mildenhall2021nerf}. Since its discovery, most methods proposed in contemporary literature are based on NeRF as a backbone. Typically, a NeRF requires several images of a scene to generate multiple views. The recent literature has drifted towards achieving a "NeRF like" reconstruction of a scene with fewer images \cite{yu2021pixelnerf}, \cite{roessle2022dense}, \cite{zhang2021ners}, \cite{niemeyer2022regnerf}, primarily depending on extraction of image based features followed by an end to end training with some 3D supervision. For instance, Pixelnerf \cite{yu2021pixelnerf} uses a CNN based feature extractor with differentiable un-projection of a feature frustum derived from the input images. Generalizable NeRF Transformers \cite{wang2022attention} completely does away with the ray tracing half of the NeRFs, replacing it with a transformer block which learns to aggregate multi-view image features. Works such as \cite{mildenhall2022nerf} have also improved the robustness of such methods to noise and quantization errors in imaging. However these methods are either restricted to per scene training of their models, as in NeRF, or require huge amount of multi-view data during the training for generalisation to unseen data. 

\subsection{Novel View Synthesis using Diffusion Models}
Diffusion based models have been proposed for object centric view synthesis; a task that is usually divided in two primary stages that includes training a 3D aware diffusion model followed by transferring the 3D consistent information to the input scene given. The score distillation sampling in \cite{poole2022dreamfusion} uses a 2D diffusion model as a prior for a parametric image generator which then is used to optimise a NeRF model for a text to 3D task. Other works such as \cite{chan2023generative}, \cite{xiang20233d} involve explicitly incorporating 3D geometric priors into diffusion models to generate 3D synthesis. While GenVS depends on evaluating models on one scene category like table or fire hydrants on a single run, zeroNVS \cite{sargent2023zeronvs} can process multiple categories for evaluation in a single model. However the visual results often are far more compromised and look blurred with zeroNVS. 

\begin{figure*}[!h]
    \centering
    \includegraphics[width=\linewidth]{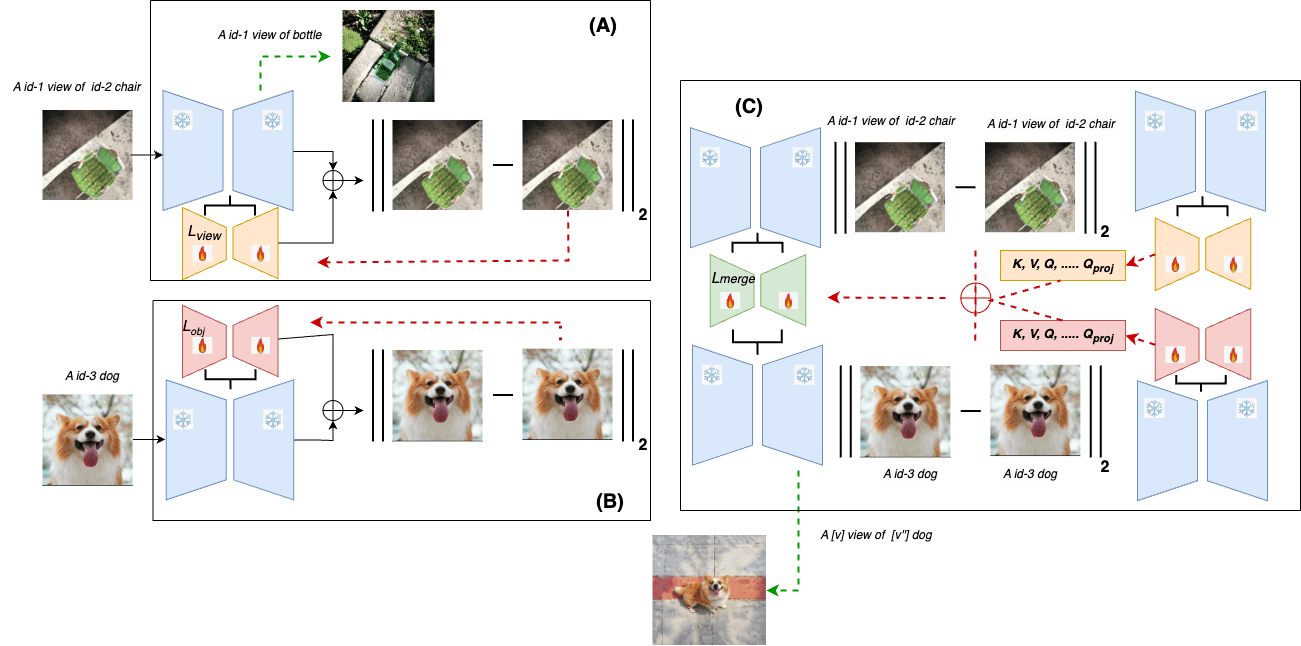}
    \caption{\textbf{Proposed approach for view transfer to unseen objects.} The blue frozen model is a SDXL \cite{podell2023sdxl} pretrained model. \textbf{(A)} We train the LoRA adaptor of the diffusion model to learn a concept of view from an image using a text prompt with two unique identifying tokens, one for the view and the other for preserving the visual concept of the original object. The green dashed line shows an inference where we simply apply the unique token of view to the specific view of a bottle. \textbf{(B)} The LoRA adaptor of the diffusion model here learns the visual content of the new images of a novel object (dog), using an unique token for it. \textbf{(C)} In this stage we merge the object and view LoRA adpators by minimizing the dot-product between the columns of the LoRA weight matrices. The red dashed line show the flow of gradient during back propagation.}
    \label{model}
\end{figure*}

Considering the fact that diffusion models are trained on vast amounts of web data \cite{rombach2022high}, much of the existing works employ finetuning strategies to harness the trained latent space of diffusion pipelines. Dreambooth \cite{ruiz2023dreambooth}, for instance, can generate high fidelity images of a specific object, using only a handful of sample images of the object in question. Such discoveries paved way to more directed and coordinated research efforts to explore the capabilities of diffusion models, strictly in the domains of novel view synthesis. Textual inversion based methods like ViewNeTI \cite{burgess2023viewpoint} takes camera viewpoint parameters, Ri, and a scene-specific token, $S_o$, to predict a latent in the CLIP output text space. They introduce neural mappers which learn to produce word embeddings for $S_o$ and camera viewpoint parameters. Nerdi \cite{deng2023nerdi} uses an image caption and word embeddings extracted from the image itself using textual inversion to feed it to the diffusion network. Most of these methods, however, rely on three dimensional priors, such as camera extrinsic and intrinsics, which leaves room to explore if diffusion models can induce a three dimensional understanding of a real scene, without explicit 3D priors.

\subsection{Unseen Domain Generalization using Low-Rank Adaptation (LoRA)}
Low Rank Adaptation \cite{hu2021lora} is an uniquely intuitive way of reducing the effective number of trainable parameters, when finetuning large scale models, such as Stable Diffusion \cite{rombach2022high}. LoRA based finetuning methods only learn a trainable matrix of a very low intrinsic rank, which makes the training and storage of these weights efficient. LoRA based finetuning poses other advantages: pre-trained models can be equipped with any set of LoRA adapted weights for different domain adaptation tasks. Methods such as \cite{shah2023ziplora} \cite{huang2023lorahub} \cite{xia2024chain}\cite{renduchintala2023tied} explore the possibility of using multiple sets of LoRA weights and efficient ways of combining these weights for the models to use. With this development as prior to our research, we are posed with a possibility that such methods, which learn finetuned concepts in a manner disjoint from the original pre-training, may also be able to learn view and object concepts. We make reasonable efforts to explore these research gaps.

\section{Methodology}
We dedicate this section to a discussion of the methodologies we adopt in our work. Our method involves low rank adapted finetuning \cite{hu2021lora} on a stable diffusion XL model \cite{podell2023sdxl} to learn object and view concepts and then eventually combine these concepts together to generate novel views. The training strategy of FSViewFusion is divided into three stages as shown in \autoref{model}. In stage-A, we finetune the base stable diffusion weights with view specific LoRA adaptors to learn the concept of view with a single image sample for the corresponding view. In the second stage-B, we train object specific LoRA adaptors, which can learn the visual attributes of the novel object from 3-4 samples of the object. Both stage-A and B LoRAs are finetuned following the Dreambooth method. Finally, in stage-C, we merge the two concepts adaptively with guidance from the previously trained LoRA adaptors. In all the three stages we keep the base stable diffusion model weights frozen updating only the key, query, value and their projections in attention modules following the LoRA \cite{hu2021lora} based finetuning literature.

\subsection{Problem Statement}

Given a set of images $\Phi_v =\{x_i\}$ where $i \in [1,N]$, $N$ being the number of samples each representing a view, and a pre-trained generative model, $D_{\theta_0}$, $\theta_0$ being the base model parameters, we finetune $D_{\theta_0}$ on $\Phi_v$ by adapting to a set of weights with a lower intrinsic rank, to that of the original weights $\theta_0$ \cite{hu2021lora}. Our assumptions are as follows: views 
exists in the latent space of generative models as high level concepts, and these can be learnt in a manner, disentangled from other concepts. The objective is to train $D_{\theta_0}$ on the high-level concept of camera view, $v$, such that the updated model low rank weights, $L_v\{D_{\theta_0}\}$, can be transferred to the unknown images of an unseen object, $\Phi_o$ with a distinct high level concept, such that the generated images, $\hat{\Phi_o}$, shares close fidelity to $\Phi_o$ while preserving the concept $v$. Our hypothesis is that the two concepts $v$ is learnt in a disentangled fashion from the object's identity. Therefore, $v$ and $o$ can eventually be merged together to generate sample image adhering to both concepts.

\subsection{View training and Object training with Low Rank Adapters}
We first address the need for using LoRAs for training. In our experience with keeping all the layers of diffusion network unfrozen for training the concept of view and object sequentially, we observed that there was catastrophic forgetting for the concept which was learnt first \cite{smith2023continual}. This is detrimental for our problem. LoRAs act as expert models that can be merged with added benefit of computational efficiency. Therefore, we train separate LoRAs for two different concepts. 

In LoRA\cite{hu2021lora} training, the weight updates $\Delta\theta$ to the base model weights $\theta$ where $\theta \in \mathbb{R}^{m \times n}$ can be decomposed into two intrinsic matrices which are lower in rank. Typically weights for layer $i$ is represented as $\theta^{\{i\}}$, but we drop the index notation for simplicity. If matrix $\Delta\theta$ is of size $m \times n$, it can be represented as a matrix multiplication of two matrices $A$ and $B$ of size $m \times r$ and $r \times n$ respectively, r being the intrinsic rank of $\Delta\theta$. Therefore, $\Delta\theta = A\cdot B$ where $A$ and $B$ are trainable. For inference, the weight matrix $\theta'$ can be obtained as $\theta + \Delta\theta = \theta + A\cdot B$. Let the pretrained stable-diffusion model $D$ be initialized with weights $\theta_0$. We finetune the model on distribution of $\phi_v$ with a unique view token for the view as well as a unique object token for the object. Thereby, the text condition to the model becomes "A \textit{unique-id-1} view of \textit{unique-id-2} [\textit{class object}]". Given this, the view specialised weight updates $\Delta\theta_v$ can be decomposed as shown in \autoref{eq-5}. Finally, the weight updates could be added to the base model weights. 
\begin{equation}\label{eq-5}
    \Delta\theta_v = A_v \cdot B_v
\end{equation}

We train the object LoRA in similar manner as the view LoRA using only one object specific \textit{unique-id} token in the prompt. Subsequently, upon training specific LoRA weights $L_v$ and $L_o$ for each concept, we merge them as laid down in \autoref{eq:lora_merge}. The only crucial difference in the two training is that the object training involves a few (3-5) training images, while the view concept training is done on only one view image.

\subsection{Low Rank Merging}\label{obj-cat}

\begin{figure*}[!h]
    \centering
    \includegraphics[width=\linewidth]{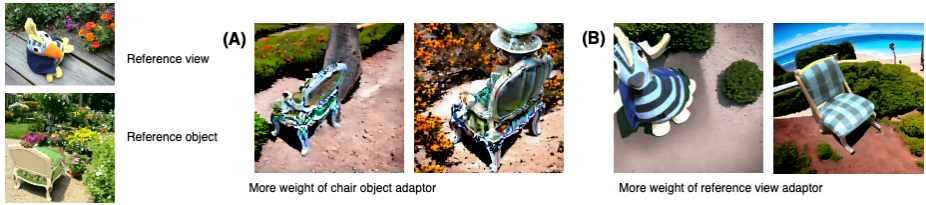}
    \caption{\textbf{The problem with linearly weighted merging.} In \textbf{part (A)} when the weight of the object adaptor is kept high we see broken reconstruction of chair. In \textbf{part (B)}, the weight of view adaptor is high resulting in concept leaks.}
    \label{leak}
\end{figure*}

 The two LoRA weights, $L_v$, $L_o$, can be merged as a linear combination of the individual weight updates. This means that the merged LoRA weights $L_{vo}$ are given as:
\begin{equation} \label{eq:lora_merge}
    L_{vo} = w_v \cdot \Delta\theta_v + w_o \cdot \Delta\theta_o = w_{vo} \cdot \Delta\theta_v + (1 - w_{vo}) \cdot \Delta\theta_o
\end{equation}
where $w_v$, $w_o$ and $w_{vo}$ are scalar weights. These weights can be tuned in order to gain control over the influence of concept learning $v$ and $o$. However, this is not well suited for our objective. The linear combination of LoRA weights results in ambiguous artifacts in image reconstructions. We observe that the identity of the object from which the view is learnt and the unseen object's identity would either superimpose with each other resulting in concept leaks or would result in broken reconstructions as shown in \autoref{leak}. 

To mitigate this issue, we adopt the style transfer merging in ZipLoRA \cite{shah2023ziplora} to transfer the concept of the view to the object. As an alternative to \autoref{eq:lora_merge},  the scalar constant, $w_{vo}$, for a better merging, can be replaced with an elementwise weight constant. To this end, the merging process becomes:
\begin{equation}
    L_{vo} = m_o \otimes \Delta\theta_o + m_v \otimes \Delta\theta_v
\end{equation}
where $m_o$ and $m_v$ represent elementwise weighing matrices for weight updates. This modification is based on the observation that LoRA matrices are sparse and the alignment of LoRA matrices must be diverse in order to have a better merging \cite{shah2023ziplora}. We observe similar trends while training the view concept as well. The idea behind this is that, the columns of weight matrices of any two independently trained LoRAs are highly aligned and merging them directly results in superimposition. Therefore for training the merged LoRA adaptor, the cosine similarity between merge vectors $m_o$ and $m_v$ is minimised, making the columns of the weights of the view and object adaptors orthogonal to each other by disentangling them. 

\section{Experiments}
In the following section, we demonstrate the efficacy of FSViewFusion in reliably reconstructing views of unseen objects from a single scene through extensive experiments. We provide the quantitative comparison between FSViewFusion and other state-of-the-art few shot NeRF reconstruction methods on a synthetic dataset in \autoref{quant} along with the qualitative results for our method on both synthetic and natural images. 

\subsection{Comparison on Datasets} \label{quant}
\textbf{Dataset and evaluation protocol}: Following ViewNeti \cite{burgess2023viewpoint} and other recent works for novel view synthesis, we evaluate FSViewFusion on the widely used DTU MVS dataset \cite{aanaes2016large}. DTU provides ground truth images corresponding to camera views of different objects. For the same camera view, ground truth images of different objects can be used for our evaluation purpose. An object can be chosen as the reference view while another can be chosen as the novel object, with the generated image evaluated based on the ground truth image of the latter for that particular camera view. We use the 15 test scenes as well as the 6 evaluation scenes of DTU which were provided in the literature \cite{burgess2023viewpoint} \cite{deng2023nerdi} \cite{yu2021pixelnerf}. In order to calculate the SSIM scores \cite{wang2004image} without having any bias due to the background, we generate the segmentation masks for both the original scenes and generated scenes using segment anything (SAM) \cite{kirillov2023segment}. The caption input to SAM is the class name for the object in the scene. For the base view training, we use a randomly selected scene from the training set of DTU dataset for fair comparison.

\noindent \textbf{Experimental Setup}: All our experiments are performed on SDXL v1.0 \cite{podell2023sdxl} base model. We follow the default settings of ZipLoRA \cite{shah2023ziplora} for finetuning our setup. Specifically, the input image is resized to $1024\times1024$. The batch size is set to $1$ for all three finetuning stages. For the view and object training stages, we finetune FSViewFusion for 1000 iterations and for the final merging stage we finetune our model for 100 iterations.  We use the default SGD optimizer with a constant learning rate scheduler with the initial learning rate set to $5e-5$. The base stable diffusion model weights and text encoders are kept frozen throughout the three training stages updating only the LoRA layers which in our case are the query, key, value and their projections in the self and cross attention modules. The rank of LoRA is set to 64 and the value of the cosine multiplier $\lambda$ is set to 0.01. The unique identifiers for each concept is chosen following the DreamBooth protocol. However, we do not use any geometric augmentations like random flipping or cropping as it changes the definition of a view.

\noindent \textbf{Baselines}: We compare FSViewFusion with two few shot state-of-the-art benchmarks PixelNerf \cite{yu2021pixelnerf} and ViewNeti \cite{burgess2023viewpoint} (both of which also perform in a single image setting). Both of these methods have prior training on the \emph{full} training set of DTU MVS dataset unlike ours which randomly selects a single scene to train the view concept which is then directly applied to train and evaluation split of DTU dataset without any pretraining. We also compare our method to NeRDi \cite{deng2023nerdi} which uses depth maps to regularize the 3D geometry. 

\noindent \textbf{Results}: 
We provide the quantitative benchmarks on three widely used metrics, SSIM \cite{wang2004image}, PSNR and LPIPS \cite{zhang2018unreasonable}. While FSViewFusion did not outperform PixelNeRF \cite{yu2021pixelnerf} by a tiny margin, our experimental setup distinctly varies from these methods. PixelNerF is directly pretrained on DTU dataset distribution whereas FSViewFusion takes a single view of one scene to learn the concept of view and 3-4 images for the new object concept itself which is not sufficient to learn the dataset's distribution. Additionally, FSViewFusion outperforms SinNeRF \cite{xu2022sinnerf}; even though SinNerf uses single image for novel view synthesis, it uses depth maps and geometric pseudo labels for regularization which is not provided in our training paradigm. Furthermore, we agree with the remarks made by \cite{deng2023nerdi} \cite{burgess2023viewpoint} that the reconstruction based metrics are not appropriate for few shot view setting as the generative models rely on hallucinating the unseen regions of the images. These metrics tend to rely on averaging over multiple views than providing a score for visually reliable views. This is an area where NeRF based methods naturally excel at. Nevertheless, as seen in row-6 in \autoref{tab:viewneti}, FSViewFusion achieves the best performance on LPIPS and PSNR, and the second best performance on SSIM, losing to PixelNeRF slightly in spite of the competitive advantage PixelNeRF has.

\begin{table}[!htbp]
\caption{Comparison for novel view synthesis on DTU dataset. The best scores are in bold and the second best are underlined.}
\centering
\label{tab:viewneti}
\resizebox{0.6\textwidth}{!}{%
\begin{tabular}{cccc}
\hline
\textbf{Methods} & \textbf{LPIPS $\downarrow$} & \textbf{SSIM $\uparrow$} & \textbf{PSNR $\uparrow$} \\ \hline
NeRF \cite{mildenhall2021nerf}            & 0.703          & 0.286         & 8.000         \\
PixelNeRF \cite{yu2021pixelnerf}       & 0.515          & \textbf{0.564}         & 16.048        \\
SinNeRF \cite{xu2022sinnerf}         & 0.525          & 0.560         & \underline{16.520}        \\
NerDi \cite{deng2023nerdi}          & 0.421          & 0.465         & 14.472        \\
ViewNeTI \cite{burgess2023viewpoint}        & \underline{0.378}          & 0.516         & 10.947        \\ 
FSViewFusion (ours) & \textbf{0.375} & \underline{0.563} & \textbf{26.587} \\
\hline
\end{tabular}%
}
\end{table}

\begin{figure*}[!h]
    \centering
    \includegraphics[width=\linewidth]{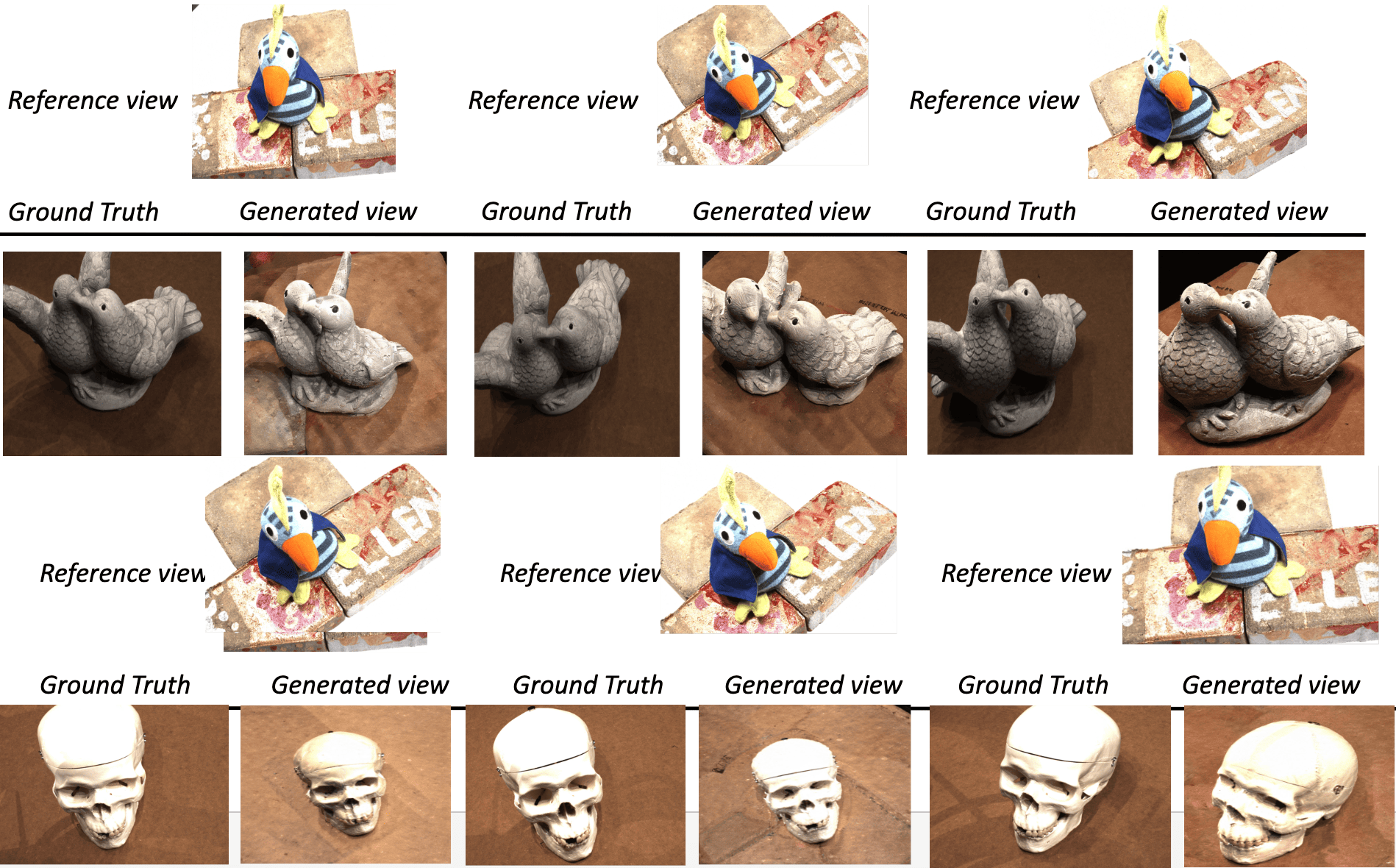}
    \caption{\textbf{Novel view synthesis on DTU MVS Dataset.} Given the reference view to train view adaptor and image samples of the novel object (statue in row-1 and skull in row-2) we compare the synthesized views with original ground truth views available. }
    \label{dtu-2}
\end{figure*}

\begin{figure*}[!h]
    \centering
    \includegraphics[width=\linewidth]{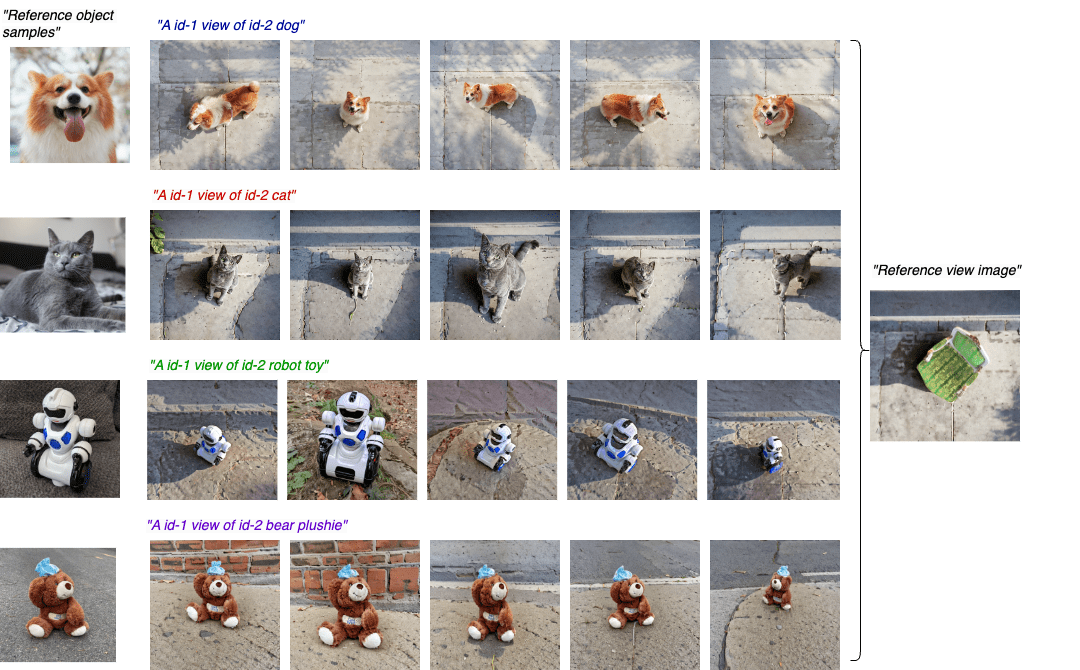}
    \caption{\textbf{Novel view synthesis on DreamBooth Dataset.} Given reference view image of a chair from top view, we reconstruct views of different objects like dog, cat, robot-toy etc. merging with given view (top) concept. FSViewFusion reliably hallucinates the top views (row-1 and row-3) irrespective of the difference of the structure of the reference view object and unseen object. As mentioned in the introduction, the degree of freedom is the object pose, if we imagine that the camera view in the chair image stays fixed while the reference object is swapped with the chair.}
    \label{dtu-1}
\end{figure*}

\noindent \textbf{Qualitative Results}: The qualitative performance of FSViewFusion on DTU MVS Dataset is given in \autoref{dtu-2}. We also provide the results on natural images from DreamBooth Dataset in \autoref{dtu-1}. For the prompts we follow the same setup as mentioned in methods, we use the class name to identify an object. For the view LoRA we use a \textit{unique-id} for the camera view and another for reference view object concepts. The object LoRA is then trained with reference images for the new object samples (3-4 images) and merged with view LoRA. As can be seen, the reconstructions look visually better for natural images in \autoref{dtu-1}. 
We attribute this to the fact that stable diffusion is pretrained on millions of natural images which allows it to hallucinate well for natural images. Unlike other methods, FSViewFusion does not learn the DTU dataset distribution for the subsequent view reconstructions. In \autoref{dtu-1} we can also see that row 1 has more reliable view reconstructions than row 4 which indicates that if the stable diffusion have seen a object in more diverse settings during its training the view reconstructions in our method become more reliable. 

\subsection{Analysis}
\begin{figure*}[!h]
    \centering
    \includegraphics[width=\linewidth]{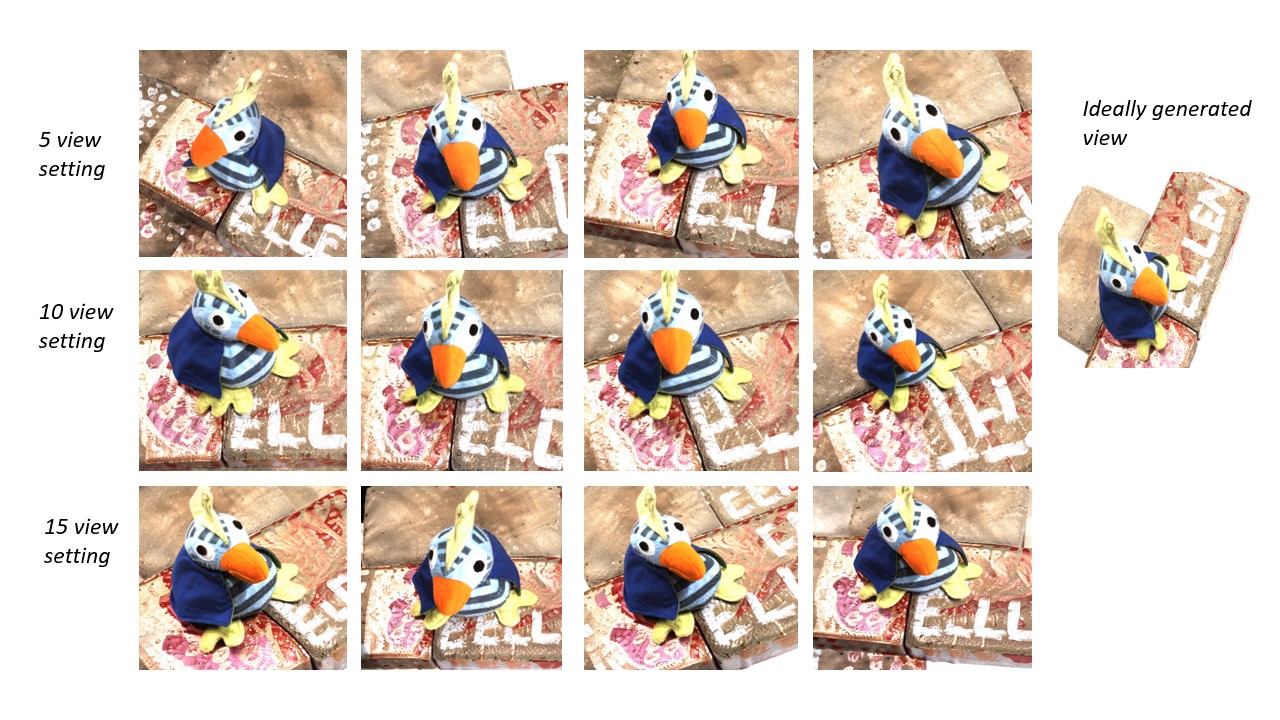}
    \caption{\textbf{Comparison of multi-view LoRAs.} We train the view LoRA adaptor with multiple views in the same model. The number of views per LoRA is varied in steps of 5,10,15. The view reconstructions in all three of the cases deviate from the reference camera view. Here, the inference was conducted on the view LoRA itself. }
    \label{A2}
\end{figure*}

\noindent \textbf{Why do we need one view per LoRA?} So far, all our experiments train one LoRA per view. In our experiments in \autoref{A2}, we attempt to train a view LoRA with multiple view concepts. In this setting, an unique identification token is assigned to each view but the unique identification token with the corresponding object concept remains the same for each view. We train the network for 5, 10 and 15 views at a time. It can be seen in \autoref{A2} that there are some outliers in all three settings. By outliers, we do not refer to the pose of the object in the image. In row 1 of \autoref{A2}, it can be seen clearly that in the first and last sample the camera view wrt to the slab on ground has changed. Even with a larger training set where we created multiple images for a specific view by inpainting with different backgrounds, such outliers remain. We conjecture that this may be because the number of training samples required increases greatly as the number of views per LoRA increases, which defeats our original few shot motivation. As a result, without loss of generality, we restrict to one view per LoRA.

\begin{figure*}[!h]
    \centering
    \includegraphics[width=\linewidth]{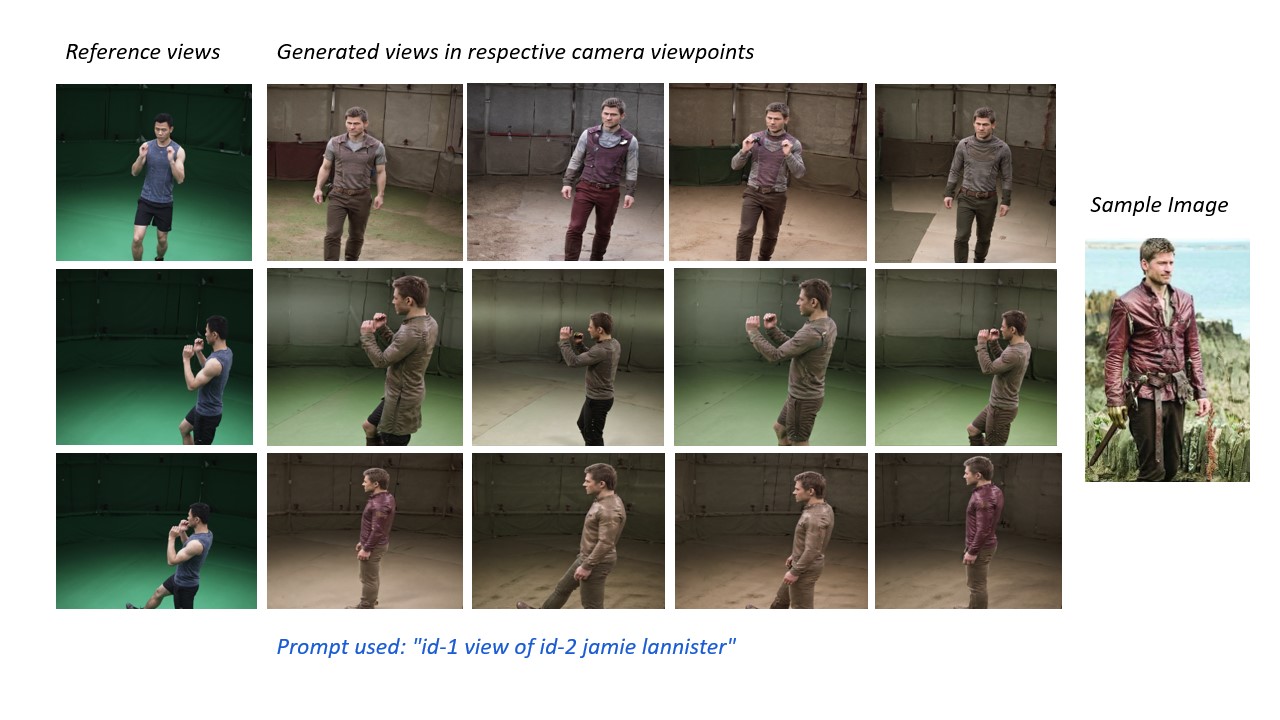}
    \caption{\textbf{Novel view synthesis on person object.} Given the reference views of athlete on the left we reconstruct a popular TV show character on the right. The object LoRA is trained with images of the character taken from web.}
    \label{A1}
\end{figure*}

\noindent \textbf{Does FSViewFusion work on complex objects?} Previously for training the view LoRA adaptor, we had selected a random view from the DTU MVS dataset, \textit{"a bird toy"} and another one from the NeRF synthetic dataset, \textit{"chair"}. Both of the objects are fairly simple to reconstruct but the question arises whether FSViewFusion can reconstruct different views of a complex object (like person) from inanimate simple objects. First, we evaluate if it is at all possible to reconstruct humans through FSViewFusion. In \autoref{A1} we use reference views of an athlete for view training and use the images of "Jamie Lannister", a character from the popular TV show Game of Thrones as the novel object. 

\begin{figure*}[!h]
    \centering
    \includegraphics[width=\linewidth]{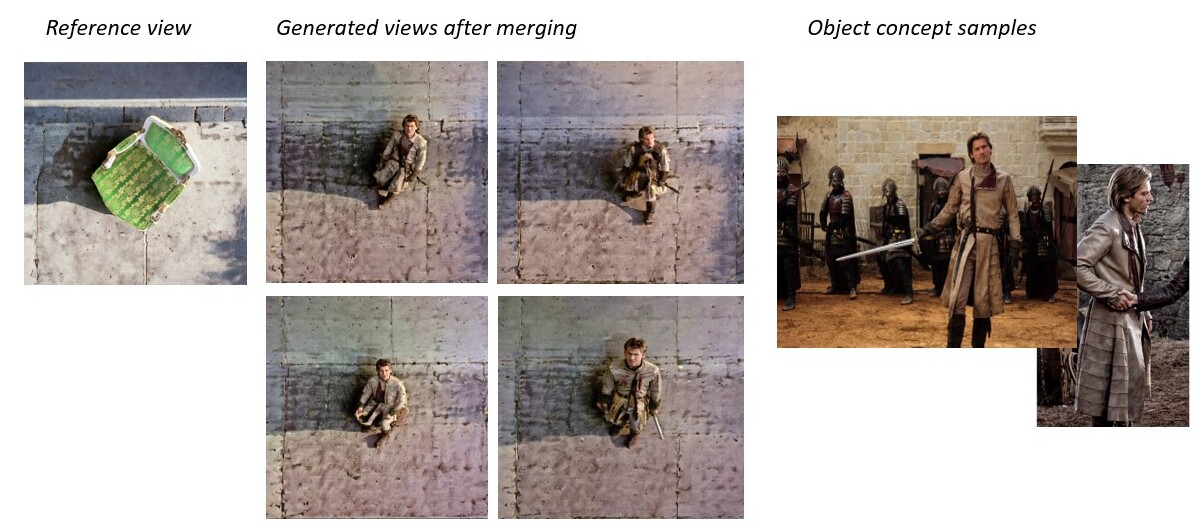}
    \caption{\textbf{Transfer of view in structurally different object.} The view LoRA trained on a chair view and object LoRA trained on a person is merged to synthesize views of the person.}
    \label{A3}
\end{figure*}

It can be seen that our method can reliably reconstruct the views of human given proper view reference images. To this end, we further evaluate our method on an extreme case of reconstructing the top views of Jamie Lannister given a chair from the top view as the reference view and only forward facing sample of the character for object training. In \autoref{A3}, it can be clearly seen that from a simple object like chair we can still reconstruct views of specific characters. Although the overall quality of reconstruction drops a little, the results are both consistent to the view and the character itself. This demonstrates the abilities of FSViewFusion for transferring the view to complex in-the-wild images.  

\begin{figure*}[!h]
    \centering
    \includegraphics[width=\linewidth]{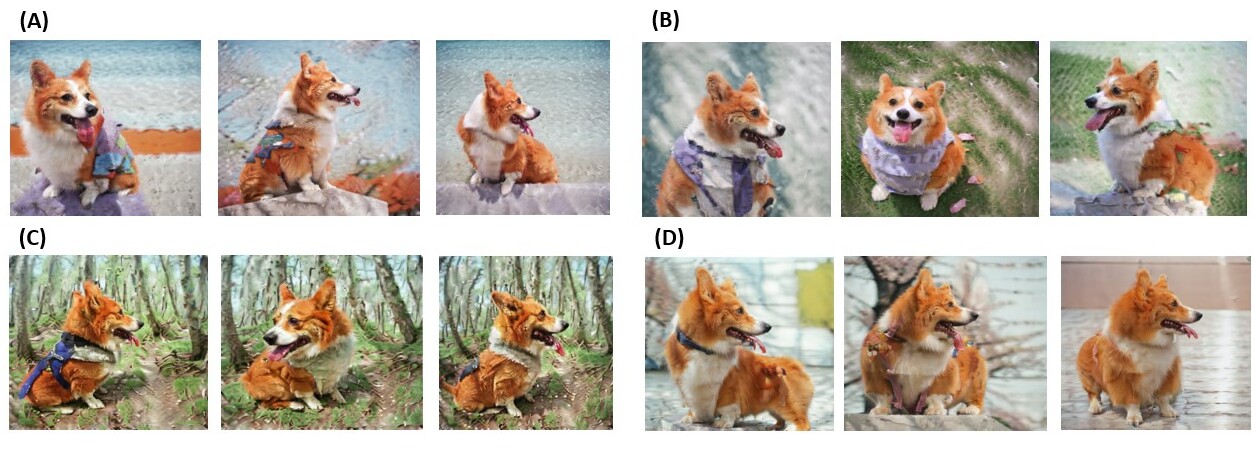}
    \caption{\textbf{Effect of backgrounds in merging the view and the object adaptors.} \textbf{Part(A)} uses beach in the background for view training. \textbf{Part(B)}, \textbf{Part(C)}, \textbf{Part(D)} uses grass, forest and table respectively. The reconstruction of views are better when bakcground has anchoring artifacts which allows the model to learn the view point of the object wrt to the objects in ground plane (e.g part (D) vs part(B).}
    \label{A5}
\end{figure*}

\noindent \textbf{Does changing the background have any effect on the view transfer?} In this section, we study the effect of background in the overall reconstructions post merging the view with object concept. We select 4 different backgrounds for the experiment; beach, forest, grass and table. In order to add the background to the reference view object we use stable diffusion as a inpainting model. We generate the mask of the object with SAM \cite{kirillov2023segment} and take an inverse of it. Then we provide the respective background prompt, e.g, "on a table" to reconstruct the background region. We now utilize these images for view training followed by merging the view with concept of a corgi dog. The results are given in \autoref{A5}. Part \textbf{(A)} shows the view reconstructions of the given beach as the background. Similarly, part \textbf{(B)}, \textbf{(C)}, \textbf{(D)} are of grass, forest and table respectively. It can be seen that when we have relatively complex backgrounds in the scene like a forest or table which have artifacts like edges of table, position and orientation of tress in forest through which the diffusion model can establish a sense of spatial relationship between the object and the background, the view (top-side in this case) reconstructions are overall faithful. Furthermore, a grass background does not have much artifacts which can give a sense of the camera view of the object and therefore the view reconstructions are variable. These observations thus support our definition of view, which suggests that camera view is estimated through other objects in ground plane. 

\section{Conclusion}
In this paper, we report an interesting finding that it seems diffusion models are capable of capturing specific viewpoint and object concept at a high level without needing any 3D prior knowledge. Harnessing the extensive coverage of a SDXL model, we conducted LoRA style learning of view and object before merging them. It appears our proposed pipeline is quite capable of disentangling and transferring the learned view to the novel object. Ablative experiments led us to believe that the view is learned via spatial relationships to the background, and the view transfer entails conceptually swapping in the novel object while keeping the camera view fixed. FSViewFusion requires no 3D priors or pretraining and could be a strong contribution to AI researchers seeking to synthesize novel views with minimal overhead as each view LoRA requires only one sample per view while the object LoRA 3-4 samples.

\noindent \textbf{Limitations}: Firstly, while FSViewFusion can reliably learn the views in a discrete manner, the continuous interpolation in view space still remains a challenge. Given two views, the intermediate view can be interpolated in NeRFs. This arise from the fact that our method does not take camera coordinates as the input. Secondly, FSViewFusion has one LoRA per view. In order to achieve interpolation continuously, multiple view information needs to be present in a single LoRA. We hope to overcome these limitations in future. 

%
%
\bibliographystyle{splncs04}
\bibliography{main}
\end{document}